\documentclass[10pt,twocolumn,letterpaper]{article}

\usepackage{iccv}
\usepackage{times}
\usepackage{epsfig}
\usepackage{graphicx}
\usepackage{amsmath}
\usepackage{amssymb}
\usepackage{booktabs}
\usepackage{bbm}
\usepackage[table]{xcolor}
\usepackage{adjustbox}
\usepackage[accsupp]{axessibility}  
\usepackage[title]{appendix}

\renewcommand{\Omega}{\varOmega}

\usepackage[pagebackref=true,breaklinks=true,letterpaper=true,colorlinks,bookmarks=false]{hyperref}
\iccvfinalcopy 

\def\iccvPaperID{1146} 

\begin{document}

\title{ObjectSDF++: Improved Object-Compositional Neural Implicit Surfaces}

\author{{\quad Qianyi Wu\textsuperscript{1} \quad Kaisiyuan Wang\textsuperscript{2}  \quad Kejie Li\textsuperscript{3}\quad Jianmin Zheng\textsuperscript{4}\quad Jianfei Cai\textsuperscript{1}} \\
{\normalsize \textsuperscript{1}Monash University}
\quad
{\normalsize \textsuperscript{2}University of Sydney}
\quad
{\normalsize \textsuperscript{3}University of Oxford}
\quad 
{\normalsize \textsuperscript{4}Nanyang Technological University}\\
{\tt\small \{qianyi.wu, jianfei.cai\}@monash.edu kaisiyuan.wang@sydney.edu.au} \\
{\tt \small kejie.li@outlook.com asjmzheng@ntu.edu.sg}
}
\maketitle

\begin{abstract}
In recent years, neural implicit surface reconstruction has emerged as a popular paradigm for multi-view 3D reconstruction. Unlike traditional multi-view stereo approaches, the neural implicit surface-based methods leverage neural networks to represent 3D scenes as signed distance functions (SDFs). However, they tend to disregard the reconstruction of individual objects within the scene, which limits their performance and practical applications. To address this issue, previous work ObjectSDF introduced a nice framework of object-composition neural implicit surfaces, which utilizes 2D instance masks to supervise  individual object SDFs. 
 In this paper,
we propose a new framework called ObjectSDF++ to overcome the limitations of ObjectSDF. First, in contrast to ObjectSDF whose performance is primarily restricted by its converted semantic field, the core
component of our model is an occlusion-aware object opacity rendering formulation that directly volume-renders object
opacity to be supervised with instance masks.
Second, we design a novel regularization term for object distinction, which can effectively mitigate the issue that ObjectSDF may result in unexpected reconstruction in  invisible regions due to the lack of constraint to prevent collisions. Our extensive experiments demonstrate that our novel framework not only produces superior object reconstruction results but also significantly improves the quality of scene reconstruction. Code and more resources can be found in \url{https://qianyiwu.github.io/objectsdf++}.
\end{abstract}

\section{Introduction}
\label{sec:intro}
Recently, the neural implicit surface representation~\cite{yariv2020multiview,yariv2021volume,wang2021neus} has revived the classical area of multi-view 3D reconstruction that constructs a 3D scene from a set of multi-view images. Unlike traditional multi-view stereo approaches, neural implicit surface reconstruction leverages neural networks to implicitly represent 3D scenes as  signed distance functions (SDFs). Due to its large modeling capability and resolution-free representation and also the wide applications of 3D reconstruction (for instance, in virtual and augmented reality and ditigal twins), this novel paradigm has brought multi-view 3D reconstruction to a new level and garnered the attention of the research community.

The success of the neural implicit surface representation should be attributed to the seminal work of Neural Radiance Field (NeRF)~\cite{mildenhall2020nerf}. NeRF revolutionarily links the neural implicit 3D radiance field with 2D images via \emph{volume rendering}, which simulates how light rays travel from the 3D scene to the 2D image plane by sampling points along each ray and accumulating their radiance. The neural implicit representation is designed as a Multi-Layer Perceptron (MLP), which takes 3D coordinates of a point as input and outputs RGB color and scene density. Since NeRF is mainly designed for view synthesis, its performance on 3D reconstruction is limited. This triggers the emergence of the neural implicit surface representation and reconstruction, which predicts an SDF value instead of a radiance field density and has proven its ability in 3D reconstruction ~\cite{yariv2020multiview,oechsle2021unisurf,yariv2021volume,wang2021neus,zhang2022iron,deng2021gram,orel2022stylesdf}.

One major limitation of most existing neural implicit surface reconstruction methods is that they basically focus on the representation of one entire scene or a single object, without paying much attention to individual objects within the scene. There have been a few attempts~\cite{yang2021objectnerf,wu2022object,kong2023vmap} that address this issue by introducing the object-compositional representation with additional guidance from 2D instance masks. In particular, ObjectSDF~\cite{wu2022object} proposes a simple and effective object-compositional neural implicit representation, where the MLP outputs individual object SDFs and the scene SDF is simply the minimum over all object SDFs. Moreover, ObjectSDF introduces a mapping function to convert each object SDF to a 3D semantic logit, which is volume-rendered to be compared with the given 2D semantic mask. 

However, ObjectSDF also has a few limitations. First, it requires the additional mapping function with a tuning parameter for constructing the vectorized semantic field from object SDFs, which introduces additional mapping noise during model optimization.
Second, it mainly emphasizes the semantic field of the front object by applying the cross entropy loss to supervise the rendered semantic label with the ground-truth semantic label, lacking the supervision for other occluded object SDFs. Third, there is no constraint between any two nearby objects to prevent collisions and overlaps. Moreover, the model convergence speed of ObjectSDF is very slow due to its heavy MLP structure.

\begin{figure*}
    \centering
    \includegraphics[width=0.98\linewidth]{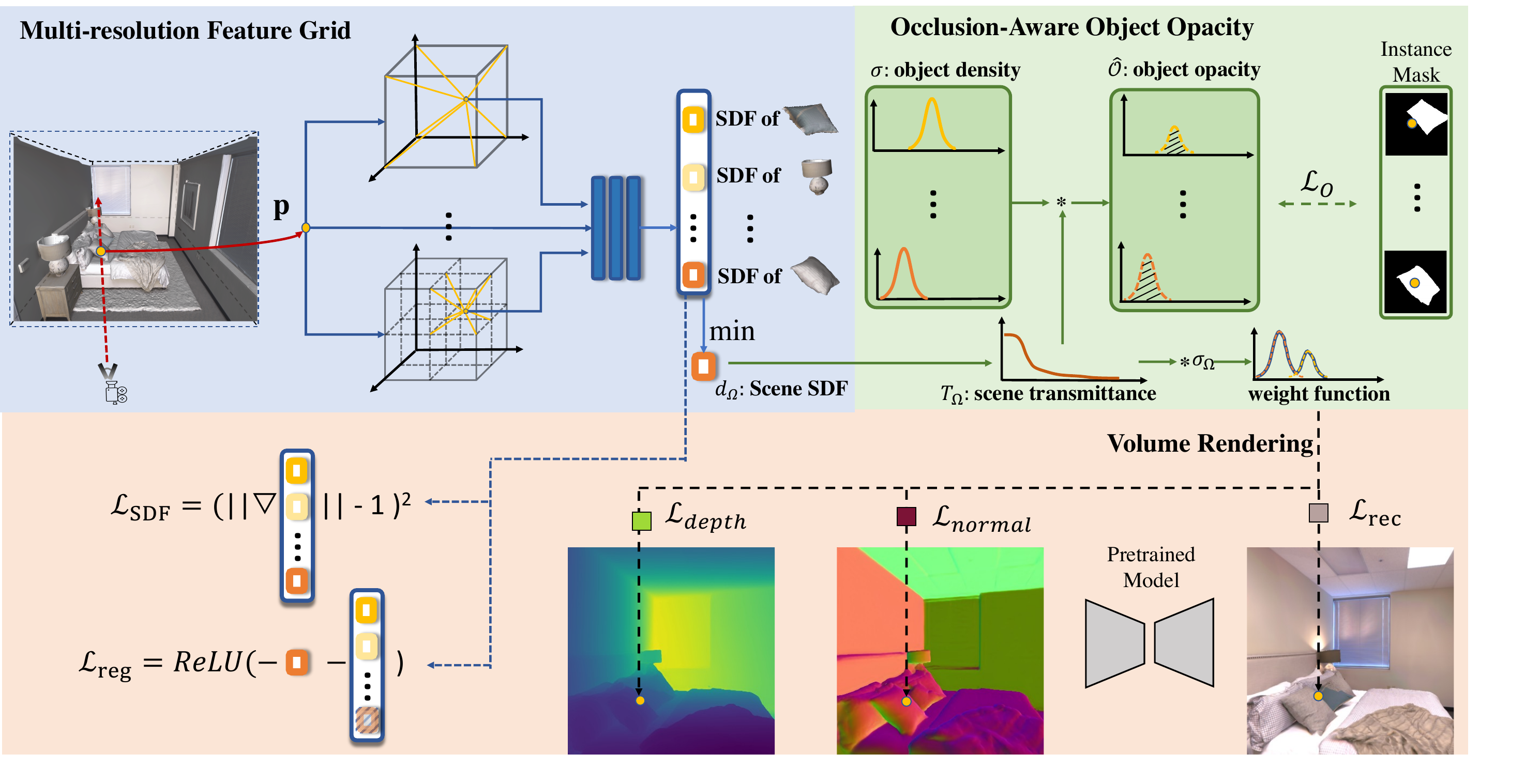}
    \caption{\textbf{Overview of our proposed framework.} Our model is able to reconstruct high-fidelity object geometry by predicting the SDF of each object inside the scene. The key to its success is the proposed occlusion-aware object opacity rendering scheme that assures the frontal object to absorb all photons during volume rendering, which helps in object surface reconstruction. A novel object distinction regularization loss is also designed to decompose objects better. To facilitate the training convergence, we use the multi-resolution feature grid and monocular geometry cues from the pre-trained model~\cite{Eftekhar_2021_ICCV} to guide the optimization of the network. 
    }
    \label{fig:pipeline}
    \vskip -0.4cm
\end{figure*}

Therefore, in this paper, we propose a new framework called \emph{ObjectSDF++} to address the  limitations of ObjectSDF. While our method is built upon ObjectSDF, it is not merely a straightforward extension of ObjectSDF. The core of our model is an occlusion-aware object opacity rendering formulation (see Fig.~\ref{fig:pipeline}), which removes the entire semantic field of ObjectSDF and directly volume-renders object opacity to predict individual object opacity masks. 
The key insight behind this is that a ground truth instance mask not only provides information about which object is in the front but also indicates that \emph{this object should absorb all the light emanating along the ray in volume rendering}. 
Our new representation gives a stronger constraint compared with the solution in ObjectSDF that only considers the rendered semantic logit close to the ground truth  label. 
Our experimental results show that such a design substantially improves the reconstruction quality of both scenes and objects.

To alleviate the problem of overgrown object shapes, we further design a physical-based regularization that penalizes an object intruding into other objects. 
The key observation is that object SDFs should be mutually exclusive.
Specifically, we notice that if one point is located inside an object with a negative SDF value of $s$, it suggests its distance to any other objects should be greater than or equal to $-s$. 
Such a simple and intuitive regularization term helps the model separate different objects and alleviate the object collision problem. In addition, 
inspired by previous works~\cite{yu2022monosdf,deng2021depth,wei2021nerfingmvs,roessle2022dense,chen2023explicit} that use geometry cues to improve neural rendering and reconstruction, we also enforce monocular geometry cues from a pre-trained model to facilitate high-quality object-compositional surface reconstruction and faster model convergence. To further speed up the training, we adopt a multi-resolution feature grid as position embedding together with a tiny MLP structure, following~\cite{mueller2022instant,yu2022monosdf}.

Overall, our main contributions are as follows.
\begin{itemize}
    \item We propose a new occlusion-aware object opacity rendering scheme to improve both scene and object reconstruction in the object-compositional neural implicit surface representation.
    \item To alleviate object collisions in the reconstruction, we introduce an object distinction regularization term, which is simple and effective.
    \item We evaluate our method on different datasets and demonstrate the effectiveness of our proposed components, not only on object-level reconstruction but also on scene-scale surface reconstruction.
\end{itemize}

\section{Related Work}
\noindent\textbf{Neural Implicit Representation}. Coordinate-based MLPs, also termed as neural implicit representation have become a powerful representation form in approximating different signals like 3D geometry. Occupancy Networks~\cite{mescheder2019occupancy} and DeepSDF\cite{ park2019deepsdf} are two pioneers which first introduce the concept of implicitly encoding objects or scenes using a neural network, despite that these studies need 3D ground-truth models. Scene Representation Networks (SRN)~\cite{sitzmann2019scene} and Neural Radiance Field (NeRF)~\cite{mildenhall2020nerf} reveal that geometry and appearance can be simultaneously learned from multiple RGB photos utilizing multi-view consistency. 
Further predictions using this implicit representation  include  semantic  labels~\cite{Zhi:etal:ICCV2021,vora2021nesf}, deformation fields~\cite{pumarola2021d,park2021hypernerf}, and high-fidelity specular reflections~\cite{verbin2021refnerf}. 

Among the existing neural implicit representation works, NeRF's learning-by-volume-rendering paradigm has received tremendous attention. It sets the basis for many subsequent efforts, including ours. On the other hand, several works~\cite{yariv2020multiview,oechsle2021unisurf,luan2021unified,yariv2021volume,wang2021neus,deng2021gram,sun2022controllable} show that rendering neural implicit surfaces, where gradients are concentrated around surface regions, can produce a high-quality 3D reconstruction. This approach often yields superior results when compared to rendering a neural radiance field. Particularly, the recent efforts~\cite{yariv2021volume,wang2021neus} combine volume rendering and neural implicit surfaces, achieving very accurate surface reconstruction. Despite the good reconstruction performance, 
all these methods mainly focus on the  representation of one entire scene or a single object. In contrast, our framework jointly represents and reconstructs both scene and its  objects in an object-compositional manner.

\noindent\textbf{Object-Compositional Neural Implicit Representation}. Decomposing a holistic neural implicit representation into several parts or  individual  object  representations can benefit many downstream applications such as scene editing, generation and understanding. 
Several attempts have been made. 
The existing methods can be categorized as category-specific or category-agnostic methods.

The category-specific methods learn the object-centric representation from many training data in the same category~\cite{epstein2022blobgan,irshad2022shapo}. For example, Ost \emph{et al.}~\cite{ost2021neural} utilize the scene graph information in rendering a dynamic street scene. Yu \emph{et al.}~\cite{yu2022unsupervised} and Guo \emph{et al.}~\cite{guo2020object} concentrate on extracting object-centric implicit representations by exploiting a large number of synthetic data. Niemeyer and Geiger~\cite{niemeyer2021giraffe} introduce GIRAFFE that conditions on latent codes to obtain object-centric NeRFs and thus represents a scene as a compositional generative neural feature field. Xu~\cite{xu2022discoscene} proposes an object-centric 3D scene generative NeRF framework DiscoScene by leveraging the 3D object location prior to achieving object-compositional generation. Although these methods can utilize the fruitful information in the dataset, they have difficulty in generalizing to unseen objects.

The category-agnostic approaches learn an object-compositional neural implicit representation without involving a large dataset for learning object priors. They rely on additional information to figure out objects in a scene. For example,    
D2Nerf~\cite{wu2022d} proposes to learn the dynamic object inside the scene from dynamic video data. Different from a dynamic video~\cite{song2022nerfplayer,tschernezki21neuraldiff} which can provide a natural decomposition between dynamic objects and the static background, learning an object-compositional representation from a static scene requires extra cue during training. The existing methods mainly adopt pixel-level guidances such as segmentation masks or neural features~\cite{tschernezki2022neural,fan2022nerf,mazur2022feature,wang2022dm,kundu2022panoptic,siddiqui2022panoptic,liu2022unsupervised,wang2023autorecon,li2023rico,xiangli2023assetfield,wang2023learning,chen2023interactive}. 
For instance, Sosuke and Eiichi~\cite{kobayashi2022distilledfeaturefields} used a powerful model like CLIP~\cite{radford2021learning} or DINO~\cite{caron2021emerging} 
to learn a neural feature field to support further editing and manipulation by feature matching. 

Compared with  the neural features,  segmentation masks can provide more accurate object information and are not difficult to obtain. In particular, ObjectNeRF~\cite{yang2021objectnerf} uses instance masks to supervise the learning of each object NeRF. To better capture the high-fidelity instance geometry, ObjectSDF~\cite{wu2022object} introduces a strong connection between each object SDF and its semantic label to obtain object-compositional neural implicit surfaces with the guidance of RGB images and their associated instance masks. However, it still suffers from several issues such as inaccurate object and scene reconstruction, slow convergence and training speed. A very recent concurrent work, vMAP~\cite{kong2023vmap}, aims to solve the object-compositional SLAM from an RGB-D video with instance masks. With the help of a depth sensor, vMAP can sample the points around the depth for fast and accurate surface reconstruction. 
In contrast, we focus on the more challenging case that has no accurate metric depth but only RGB multiview images and instance masks, same as the setting of ObjectSDF. Our method is built upon ObjectSDF but resolves its limitations.

\section{Methodology}

Given a set of $N$ posed RGB images  $\mathcal{I}=\{I_1,\dots, I_N\}$ with the associated instance-segmentation mask $\mathcal{S}=\{S_1,\dots, S_N\}$, our goal is to reconstruct the scene geometry in an object-compositional neural implicit representation which describes not only the entire scene geometry but also each individual instance's geometry. 
We propose a new framework called {\em ObjectSDF++} for this purpose, which includes several modules as shown in Fig.~\ref{fig:pipeline}. The technical details of the core modules are explained in this section.



\subsection{Preliminary}\label{sec:pre}
\noindent\textbf{Volume Rendering of Scene Density.} The key to learning the neural implicit representation from multi-view images is volume rendering~\cite{kajiya1984ray}. Considering a ray $\mathbf{r}(v) = \mathbf{o} + v\mathbf{d}$ emanated from a camera position $\mathbf{o}$ in the direction of $\mathbf{d}$, the volume rendering is essentially about the integrated light radiance along the ray. There are two crucial factors in the volume rendering: scene density $\sigma(v)$ and scene radiance $\mathbf{c}(v)$ on each 3D point along a ray. An important quantity used in the volume rendering computation is the scene transmittance function $T(v)=\exp(-\int_{v_n}^v\sigma(\mathbf{r}(u))du)$, measuring the energy loss during the ray traversing. Thus, the scene opacity $O$ can be defined as the complement probability of $T$, \emph{i.e.}, $O(v) = 1 - T(v)$. 
In this way, a target pixel color of volume rendering can be calculated as the integral of the probability density function of scene opacity $\frac{dO(v)}{dv}=T(v)\sigma(\mathbf{r}(v))$~\cite{tagliasacchi2022volume} and 
radiance $\mathbf{c}(\mathbf{r}(v))$ along the ray from near bound $v_n$ to far bound $v_f$:
\begin{equation}  \label{eq:nerf}
\hat{C}(\mathbf{r}) =\int_{v_n}^{v_f}T(v)\sigma(\mathbf{r}(v))\mathbf{c}(\mathbf{r}(v))dv. 
\end{equation}
This integral is approximated using numerical quadrature.  

\noindent\textbf{SDF-based Neural Implicit Surface Representation.} SDF representation is often used to directly characterize the geometry at the surface. 
Specifically, given a scene $\mathcal{\Omega}\subset \mathbb{R}^3$, $\mathcal{M} = \partial\mathcal{\Omega}$ is the boundary surface. The SDF $d_{\mathcal{\Omega}}$ is defined as the signed distance from point $\mathbf{p}$ to the boundary $\mathcal{M}$:
\begin{equation}
    d_{\mathcal{\Omega}}(\mathbf{p}) =  (-1)^{\mathbbm{1}_{\mathcal{\Omega}}(\mathbf{p})}\min_{\mathbf{y}\in \mathcal{M}} || \mathbf{p} - \mathbf{y}||_{2},
    \label{eq:sdf}
\end{equation}
where $\mathbbm{1}_{\mathcal{\Omega}}(\mathbf{p})$ is the indicator function denoting whether $\mathbf{p}$ belongs to the scene $\mathcal{\Omega}$ or not. If the point is outside the scene, $\mathbbm{1}_{\mathcal{\Omega}}(\mathbf{p})$ returns $0$; otherwise returns $1$. The standard 
$l_2$-norm is commonly used to compute the distance. 

To learn the SDF from multiview images without the need for ground truth SDF values, pioneer works~\cite{wang2021neus,yariv2021volume} combine SDF with neural implicit function and volume rendering to get supervision from RGB images. The common idea is to replace the NeRF volume density output $\sigma(\mathbf{p})$ with a transformation function of the SDF value $d_{\mathcal{\Omega}} (\mathbf{p})$. Specifically, in~\cite{yariv2021volume}, the density $\sigma(\mathbf{p})$ is converted from a specific  tractable transformation: 
\begin{equation}
    \sigma(\mathbf{p}) = \left\{
    \begin{aligned}
    & \frac{1}{2\beta} \exp{(\frac{-d_{\mathcal{\Omega}}(\mathbf{p})}{\beta})}   & \text{if}~ d_{\mathcal{\Omega}}(\mathbf{p}) \geq 0 \\
    & \frac{1}{\beta} - \frac{1}{2\beta} \exp{(\frac{d_{\mathcal{\Omega}}(\mathbf{p})}{\beta})} & \text{if}~ d_{\mathcal{\Omega}}(\mathbf{p}) < 0
    \end{aligned}
    \right.
    \label{eq:densityconvert}
\end{equation}
where $\beta$ is a learnable parameter to decide the sharpness of the surface density. This design implicitly assumes that the object's surface is solid. The neural implicit surface is supervised by RGB reconstruction loss and an Ekional regularization loss~\cite{gropp2020implicit} for the SDF field. 

\noindent\textbf{Modeling an object-compositional scene geometry.}
To represent each individual instance geometry inside the scene, the previous method ObjectSDF~\cite{wu2022object} predicts the SDF of each object. Suppose that one scene $\Omega$ is a composition of $K$ different objects $\{\mathcal{O}_i\in \mathbb{R}^3|i=1,\dots,K\}$, we have $\Omega=\bigcup\limits_{i=1}^{k}\mathcal{O}_i$. Then,  the scene SDF can be calculated as the minimum of all object SDFs:
\begin{equation}
d_{\Omega}(\mathbf{p})=\min_{i=1\dots k}d_{\mathcal{O}_i}(\mathbf{p}).
\end{equation}With the scene SDF, volume rendering can be applied to render corresponding RGB images, normal maps, \emph{etc}. 

\subsection{Occlusion-Aware Object Opacity Rendering}
In order to learn the individual object geometry from the supervision of instance segmentation masks, the  state-of-the-art approach  ObjectSDF~\cite{wu2022object} adopts a Sigmoid-type transition function to convert each object SDF to a 3D semantic logit, which only changes rapidly around the object boundary. 
With the help of volume rendering over the transformed semantic field, the model will push the surface of the corresponding object to the front to meet the guidance from the 2D instance~\cite{wu2022object}.


\begin{figure}
    \centering
    \includegraphics[width=\linewidth]{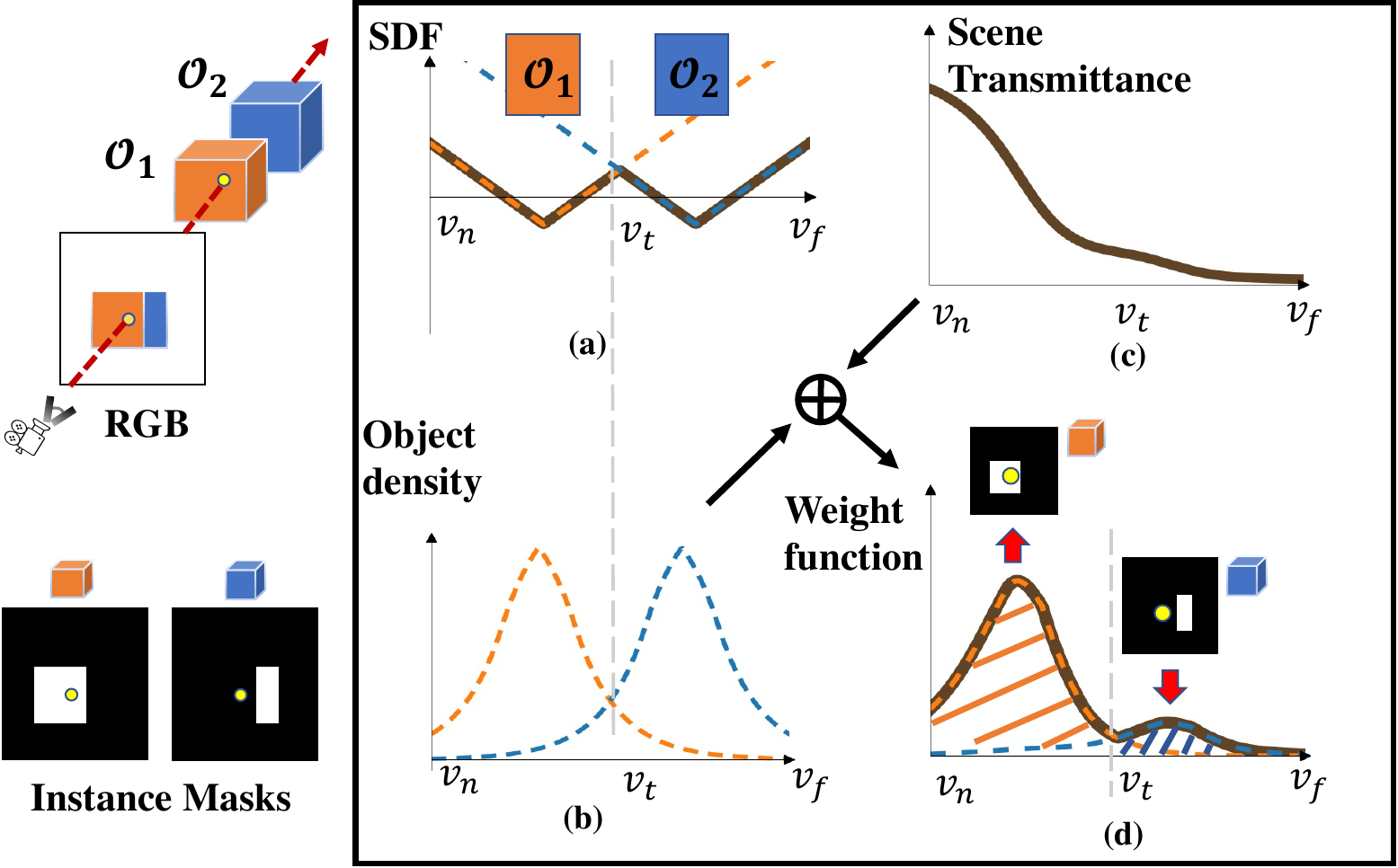}
    \caption{\textbf{Occlusion-Aware Object Opacity Rendering}. We show a toy example of occlusion-aware object opacity rendering. The opacity is essentially the integration of object density and scene transmittance. For different objects, the integration interval will differ according to its range. We propose an approximated form of object opacity rendering, which can be supervised by the pixel value in the ground truth instance mask.
    }
    \label{fig:occulsion_opacity}
\end{figure}

However, such a design has a few limitations. First, it mainly focuses on the semantic field of the front object by employing the cross entropy loss to oversee the predicted semantic label with the semantic logit of reality, which fails to provide guidance for the SDFs of obstructed objects.
Second, the softmax operation across different rendered semantic labels often has energy leaking, which cannot exactly match the one-hot ground-truth label. Third, it requires an additional mapping function to convert object SDFs to semantic logits with a tuning hyperparameter. 

To tackle these limitations, we propose to remove the need to build a vectorized semantic field and directly volume-render the opacity of each object.
Specifically, consider that when a frontal object occludes another object, the frontal one should block \emph{all} the photons emanating along the inverse ray. In terms of volume rendering, the object opacity of the frontal object should be $1$ while that of the occluded object should be $0$. For example, as shown in Fig~\ref{fig:occulsion_opacity}, the ray casting from left to right crosses the two color cubes  $\mathcal{O}_{1}$ and $\mathcal{O}_2$. If we inspect the selected pixel in the instance mask, we are not able to see the blue cube from a ray casting through the selected pixel (bottom-left of Fig.~\ref{fig:occulsion_opacity}). It means that when the photons travel through the inverse ray casting from the camera, the transmittance should become zero after they cross the \emph{first} object. As all the objects decompose the entire scene, they naturally form a Voronoi diagram~\cite{aurenhammer2013voronoi} based on individual object SDF. The light energy absorbed by the first object can be approximated by the integration of the weight function from near $v_n$ to the SDF turning point $v_t$, (dash line in Fig~\ref{fig:occulsion_opacity}(a), where the minimum SDF field changes from $\mathcal{O}_1$ to $\mathcal{O}_2$). Specifically, the object opacity for the first object $\mathcal{O}_1$ 
can be expressed as:
\begin{equation}
\begin{aligned}
    O_{\mathcal{O}_1}(\mathbf{r}) &= \int_{v_n}^{v_t}T_{\Omega}(v)\sigma_{\Omega}(\mathbf{r}(v))dv \\
    &= \int_{v_n}^{v_t}T_{\Omega}(v)\sigma_{\mathcal{O}_1}(\mathbf{r}(v))dv \\
\end{aligned}\label{eq:occ_opa_eq}
\end{equation}
where $T_{\Omega}, \sigma_{\Omega}$ are the scene transmittance and scene density function. Because in 
$[v_n, v_t]$, the scene SDF is dominated by the SDF of $\mathcal{O}_1$, resulting in 
$\sigma_{\Omega}=\sigma_{\mathcal{O}_1}$. 

This rendering equation can also be used to render the second object opacity by changing the integration interval, \emph{i.e,} $O_{\mathcal{O}_2}(\mathbf{r}) = \int_{v_t}^{v_f}T_{\Omega}(v)\sigma_{\mathcal{O}_2}(\mathbf{r}(v))dv$. 
Thanks to the consideration of scene transmittance in the integration, such an object opacity rendering is occlusion-aware.
%
However, it is hard to find the scene SDF turning points during training. On the other hand, consider the following relationship
\begin{equation}
\begin{aligned}
    {O}_{\mathcal{O}_2}(\mathbf{r}) &=\int_{v_t}^{v_f}T_{\Omega}(v)\sigma_{\mathcal{O}_2}(\mathbf{r}(v))dv \\
    &\leq \int_{v_n}^{v_f}T_{\Omega}(v)\sigma_{\mathcal{O}_2}(\mathbf{r}(v))dv,
\end{aligned} \label{eq:oo2}
\end{equation}
where the non-negative of transmittance and density guarantee this inequality. This motivates us to approximate the first integration in \eqref{eq:oo2} by the second one since $\sigma_{\mathcal{O}_2}$ is expected to be close to 0 in $[v_n, v_t]$, and similarly extend the  integration in \eqref{eq:occ_opa_eq} to $[v_n, v_f]$ since $T_{\Omega}$ is expected to be close to 0 in $[v_t, v_f]$.  
In this way, we reach 
a general   form of approximated occlusion-aware object opacity rendering:
\begin{equation}
    \hat{O}_{\mathcal{O}_i}(\mathbf{r}) =\int_{v_n}^{v_f}T_{\Omega}(v)\sigma_{\mathcal{O}_i}(\mathbf{r}(v))dv, 
    i=1,\dots, K .
\end{equation}
Then, we adopt each instance segmentation mask as the opacity supervision to guide the learning of each object surface. The loss function is defined as:
\begin{equation}
    \mathcal{L}_{O} = \mathbb
{E}_{\mathbf{r}\in \mathcal{R}}[\frac{1}{K}\sum_{i=1\dots K}\|\hat{O}_{\mathcal{O}_i}(\mathbf{r}) - O_{\mathcal{O}_i}(\mathbf{r})\|]
\end{equation}
where $\mathcal{R}$ denotes the set of rays in a minibatch and $O_{\mathcal{O}_i}(\mathbf{r})$ is the GT object opacity from the instance mask. Further discussion and detailed information can be found in appendix Sec.~\ref{sec:dis}.

\subsection{Object Distinction Regularization Term}
Although the occlusion-aware object opacity rendering equation describes the proper light behavior during the volume rendering, there is still a lack of enough constraint to regulate individual object geometries  when the transmittance approaches zero. In fact, the 2D guidance of multiview images and instance segmentation, and the geometry guidance like depth or normal map, can only guide the visible regions of a neural radiance field. The information in invisible regions could be  arbitrary since they have no influence on the loss metric between the rendered results and the 2D guidance. It could make an object  overgrow in an invisible region just like an underwater iceberg. 

To alleviate this issue, we enforce that each object inside the scene should be distinct and there should be no overlap between any two objects. In particular, suppose there are $K$ objects in a scene, for any input point coordinates $\mathbf{p}$, our model will predict the $K$ SDFs for the $K$ object surfaces. If the point is located inside the object $\mathcal{O}_1$ with a negative SDF value $d_{\mathcal{O}_1}$, one important property for the remaining objects $\{d_{\mathcal{O}_2}, d_{\mathcal{O}_3}, \dots, d_{\mathcal{O}_K}\}$ is that their SDFs need to be positive and should not less than $-d_{\mathcal{O}_1}$. In other words, it also means for the rest of the objects, their SDFs should be large than the scene SDF $-d_{\Omega}$ (\emph{i.e,} $d_{\mathcal{O}_i}\geq-d_\Omega$,  $\forall d_{\mathcal{O}_i}\neq d_{\Omega}$).
Therefore, for the SDF vector prediction of one point $\mathbf{p}$, we propose to minimize the following object distinction regularization term:
\begin{equation}
    \mathcal{L}_{reg} = \mathbb{E}_{\mathbf{p}}[\sum_{d_{\mathcal{O}_i}(\mathbf{p})\neq d_{\Omega}(\mathbf{p})}\text{ReLU}(-d_{\mathcal{O}_i}(\mathbf{p})-d_{\Omega}(\mathbf{p}))]
\end{equation}
We also find that such a requirement also holds for any point located outside of all objects. In this case, $d_\Omega>0$ and for each object SDF $d_{\mathcal{O}_i} \geq d_\Omega > -d_\Omega$. Hence, we random sample points in the entire 3D space to apply this regularization term.

\begin{figure*}[t]
    \centering
    \includegraphics[width=\linewidth]{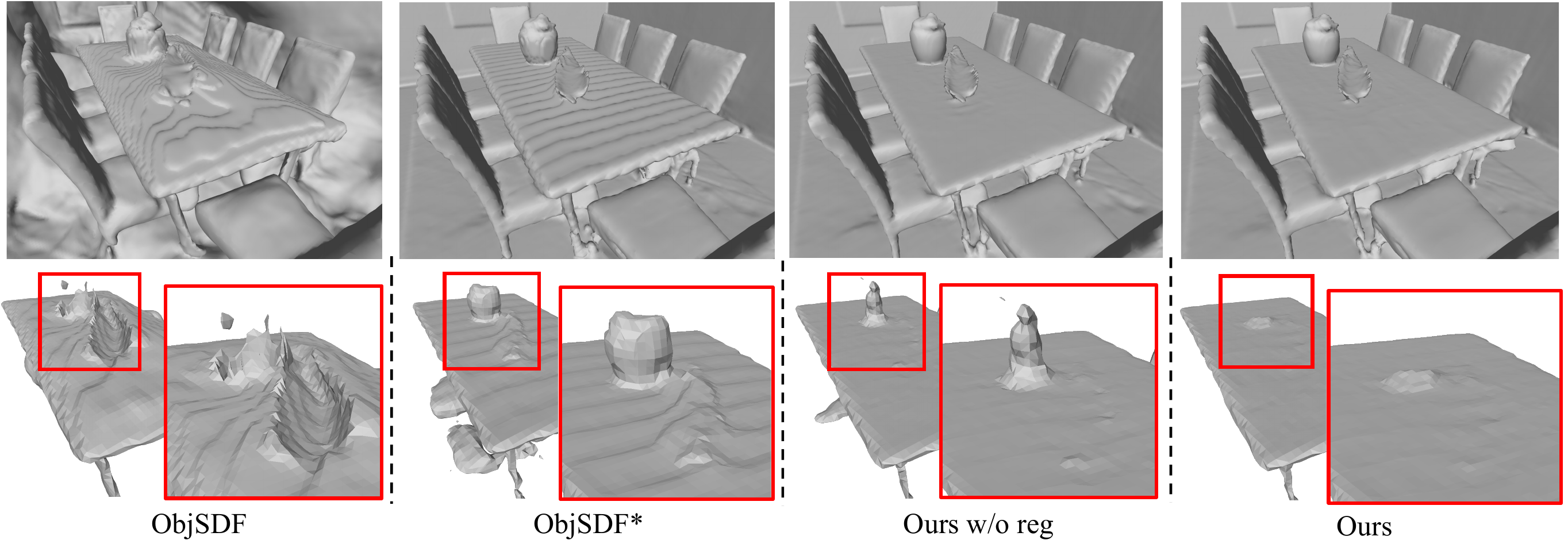}
    \caption{\textbf{Qualitative improvements of our proposed components.} The first row is the scene reconstruction results and the second row is the reconstruction of a single object (the dining table). Our model aims to improve both scene and object reconstruction in an object-compositional way. We can see that the occlusion-aware object opacity can significantly improve the surface reconstruction and the object distinction regularization term can further help to reduce the artifacts in the invisible region.
    }
    \label{fig:ablation}
\end{figure*}

\begin{table*}[t]
  \centering
    \begin{adjustbox}{width=\textwidth}
    \small
  \begin{tabular}{l|ccc||cc|cc}
    \toprule
      & \multicolumn{3}{c||}{Model Components} & \multicolumn{2}{c|}{Scene Reconstruction} & \multicolumn{2}{c}{Object Reconstruction} \\
        Method  & Object Guidance & Mono Cue& Regularizer & Chamfer-L1$\downarrow$ & F-score $\uparrow$ & Chamfer-L1$\downarrow$ & F-score $\uparrow$\\
    \midrule
     ObjSDF~\cite{wu2022object} & Semantic &   & & 22.8$\pm$14.8 & 25.74 $\pm$5.70& 7.05$\pm$0.40 & 59.91$\pm$6.67  \\
     \midrule
     ObjSDF* & Semantic & \checkmark& & 4.14$\pm$1.53 &78.34$\pm$15.34& 4.65$\pm$0.11& 74.06$\pm$3.12\\
     Ours w/o reg& Occlusion Opacity & \checkmark& & 3.60$\pm$ 1.13 & 85.59$\pm$ 8.12& 3.78$\pm$ 0.09& 79.51$\pm$ 2.92\\
     Ours & Occlusion Opacity & \checkmark&\checkmark & 3.58$\pm$1.08 & 85.69 $\pm$6.55& 3.74$\pm$ 0.09& 80.10 $\pm$ 2.63 \\ 
    \bottomrule
  \end{tabular}
    \end{adjustbox}
  \caption{\textbf{The quantitative average results from 8 Replica scenes evaluated on scene and object reconstruction}. We compare our results with vanilla ObjectSDF~\cite{wu2022object} and its improved version, ObjSDF* (adding monocular geometry cue).} 
  \label{tbl:com_replica}
  \vspace{-6mm}
\end{table*}

\subsection{Model Training}
The slow convergence of implicit neural surfaces is a common issue in most of the existing methods. The inherent reason is the heavy structure of MLP as a network backbone. To accelerate the convergence speed, inspired by the current efficient neural implicit represent work~\cite{mueller2022instant,wu2022voxurf,yu2022monosdf}, we take a multi-resolution feature grid (multi-res grid) as position embedding together with a tiny MLP to boost the model convergence and the inference speed.

To optimize the network of our object-compositional neural implicit surface, we first minimize the image reconstruction loss between the predicted color $\hat{C}(\mathbf{r})$ and the ground truth color $C(\mathbf{r})$:
\begin{equation}
    \mathcal{L}_{rec} = \sum_{\mathbf{r}\in\mathcal{R}}\|\hat{C}(\mathbf{r})-C(\mathbf{r})\|_1
\end{equation}
where $\mathcal{R}$ denotes the set of sampling rays in a minibatch. Following~\cite{gropp2020implicit,yariv2021volume,wu2022object}, we add an Eiknoal term on the sampled points to regularize object SDFs and scene SDF:
\begin{equation}
    \mathcal{L}_{SDF} = \sum_{i=1}^{k}\mathbb{E}_{d_{\mathcal{O}_i}}(|| \nabla d_{\mathcal{O}_i}(\mathbf{p})|| - 1)^2 + \mathbb{E}_{d_{\Omega}}(|| \nabla d_{\Omega}(\mathbf{p})|| - 1)^2. 
\end{equation}
Moreover, recent approaches~\cite{yu2022monosdf,wang2022neuris,guo2022manhattan} have proven that surface reconstruction could be significantly improved by incorporating geometry guidance such as depth and surface normal. Thus, we adopt the design from~\cite{yu2022monosdf} to enforce the similarity between the rendered depth/normal map and the estimated monocular depth/normal from a pretrained model Omnidata~\cite{Eftekhar_2021_ICCV}, which is termed as the monocular cue loss $\mathcal{L}_{mono}=\lambda_{depth}\mathcal{L}_{depth}+\lambda_{normal}\mathcal{L}_{normal}$. Finally, the overall training loss of ObjectSDF++:
\begin{equation} \label{eq:overall}
\mathcal{L}=\mathcal{L}_{rec}+\mathcal{L}_{O}+\mathcal{L}_{mono}+\lambda_1\mathcal{L}_{SDF}+\lambda_2\mathcal{L}_{reg}.
\end{equation}
\section{Experiment}
\begin{figure*}
    \centering
    \includegraphics[width=\linewidth]{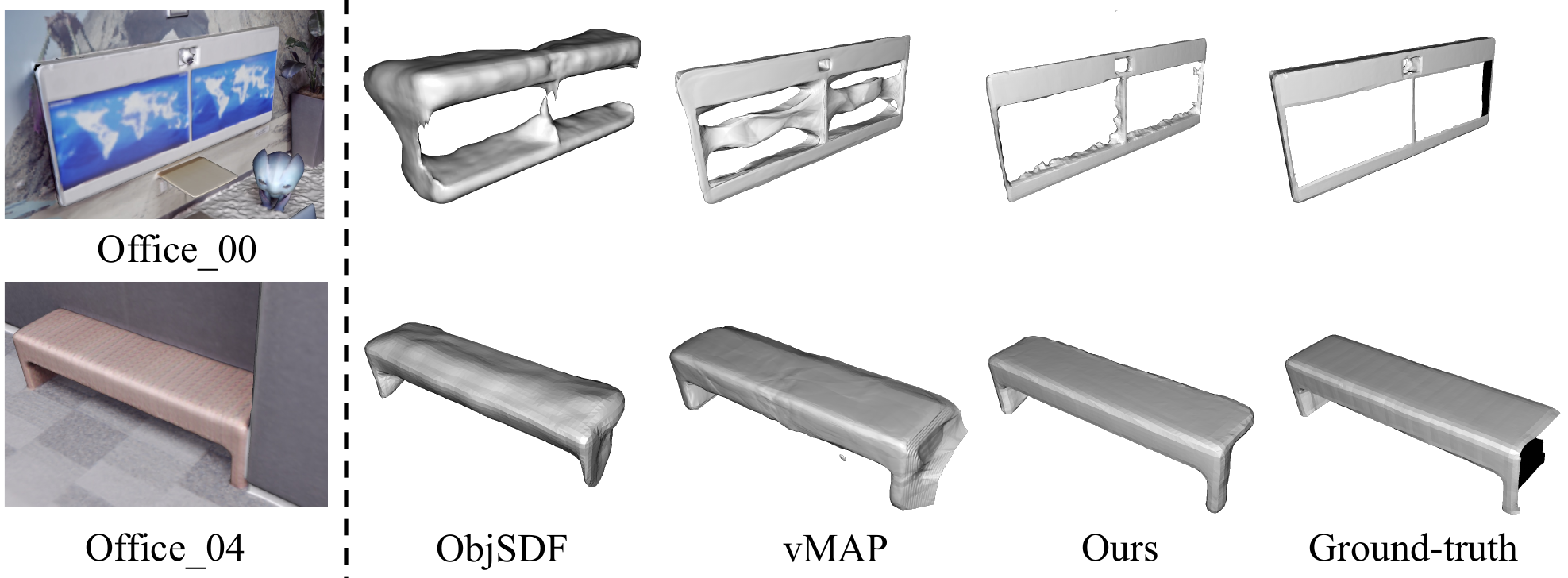}
    \caption{\textbf{Qualitative comparisons of object reconstructions on Replica by our method and two SOTA approaches~\cite{wu2022object,kong2023vmap}}. Our model can produce fewer artifacts in the invisible region due to the guidance of the regularization term and occlusion-aware object opacity rendering. All meshes of vMAP are provided by the original authors. 
    }
    \label{fig:replica_obj_comp}
    \vspace{-3mm}
\end{figure*}
Our proposed ObjectSDF++ aims to create a neural implicit representation for a high-fidelity object-compositional surface reconstruction that can be beneficial for both object and scene reconstruction. To assess the effectiveness of our approach, we conducted experiments on two datasets:  one simulated and one real-world dataset. To evaluate  both scene and object reconstruction, we compare ours with the SOTA methods for object-compositional surface reconstruction. We further compare our approach with many SOTA methods for neural surface reconstruction on large-scale real-world scenes to examine the benefit of object-compositional surface reconstruction.

\noindent\textbf{Implementation details.} We implement our method in Pytorch and apply the Adam optimizer with a learning rate of $5e-4$ for the tiny MLP part (2 layers with 256 channels for the SDF prediction and 4 layers with 256 channels for the color prediction) and $1e-2$ for the multi-res feature grid. The hyperparameters $\lambda_{depth}, \lambda_{normal}, \lambda_1, \lambda_2$ in \eqref{eq:overall} are set as $0.1, 0.05, 0.1, 0.5$ respectively. For each minibatch during training, we sample $1024$ rays and apply the error-bounded sampling algorithm~\cite{yariv2021volume} to decide the sampling points along a ray. Our method can be trained in a single 12 GB GTX 2080Ti GPU. The initial value of $\beta$ in ~\eqref{eq:densityconvert} is $0.1$. More details can be found in the appendix Sec.~\ref{sec:app_imp}.

\noindent\textbf{Dataset and metrics.}
We conduct our experiments on two datasets: a) Replica~\cite{replica19arxiv} and b) ScanNet~\cite{dai2017scannet}. Replica is a synthesized dataset with accurate camera poses and clear object masks. 
Following the setting in 
~\cite{Zhu2022CVPR,yu2022monosdf,kong2023vmap}, we use 8 scenes from Replica for experiments. For quantitative evaluation, we measure Chamfer Distance together with F-score with a threshold of $5$cm on Replica\footnote{Details about metric definition are provided in appendix Sec.~\ref{sec:metric} .}. 
ScanNet is a real-world dataset captured by an RGB-D camera. It contains 2D instance segmentation masks with RGB images. Following the setting~\cite{guo2022manhattan,yu2022monosdf,Zhu2022CVPR}, we select 4 scenes from ScanNet for experiments and report the accuracy, completeness, Chamfer Distance, precision, recall and F-score.

\noindent\textbf{Baseline Methods.}
We mainly compare our method with the following representative works in the realm of object-compositional neural implicit surface reconstruction.

\textbf{ObjectSDF~\cite{wu2022object}.} ObjectSDF is the first to use the neural implicit representation for object-compositional surface reconstruction. Its key idea is to predict individual object SDFs and convert each of them to a semantic label with a pre-defined function. The converted 3D semantic labels will accumulate along the ray by volume rendering and be supervised with the 2D semantic map. 

\textbf{ObjectSDF*.} Considering our method was built upon ObjectSDF with several improved designs. To better verify the effectiveness, we also create another baseline called \textbf{ObjectSDF*}, where we simply
add the monocular guidance and multi-res feature grid into the vanilla ObjectSDF to  improve the  reconstruction and the model convergence. 

\textbf{vMAP}~\cite{kong2023vmap}. vMAP is the latest concurrent work, which uses RGB-D images for an object-level dense SLAM system. With accurate depth information, it can sample points near the surface to handle the occlusion intuitively. We also compare our object reconstruction results with it.

\begin{table*}[t]
  \centering
  \begin{tabular}{lccccccc}
    \toprule
    ~ & Accuracy $\downarrow$& Completeness $\downarrow$& Chamfer-L1 $\downarrow$& Precision $\uparrow$& Recall $\uparrow$& F-Score $\uparrow$ \\
    \midrule
     UNISURF~\cite{oechsle2021unisurf}&0.554&0.164&0.359&0.212&0.362&0.267\\
Neus~\cite{wang2021neus}  & 0.179 & 0.208 & 0.194 & 0.313 & 0.275 & 0.291 \\
     VolSDF~\cite{yariv2021volume}  & 0.414 & 0.120 & 0.267 & 0.321 & 0.394 & 0.346 \\
     Manhattan-SDF~\cite{guo2022manhattan}  &0.072&0.068&0.070&0.621&0.586&0.602\\
     NeuRIS~\cite{wang2022neuris} &0.050&0.049&0.050&0.717&0.669&0.692\\
     MonoSDF (Multi-Res Grids)~\cite{yu2022monosdf}& 0.072 & 0.057& 0.064& 0.660&0.601&0.626\\
     Ours (Multi-Res Grids) & 0.047&\cellcolor{red!25}0.045&\cellcolor{yellow!25}0.046&0.749 &\cellcolor{red!25}0.707&0.726 \\
     \midrule 
     MonoSDF (MLP)~\cite{yu2022monosdf} & \cellcolor{red!25}0.035 & \cellcolor{yellow!25}0.048 & \cellcolor{red!25}0.042 & \cellcolor{red!25}0.799 & 0.681 & \cellcolor{yellow!25}0.733 \\
     Ours (MLP) & \cellcolor{yellow!25}0.039 & \cellcolor{red!25}0.045 & \cellcolor{red!25}0.042 & \cellcolor{yellow!25}0.781 & \cellcolor{yellow!25}0.706 & \cellcolor{red!25}0.740 \\
    \bottomrule
  \end{tabular}
    \caption{\textbf{The quantitative results of the scene reconstruction  on ScanNet}. We compare our method against various recent neural implicit surface reconstruction methods. The top-2  results are highlighted in red and yellow, respectively.}
  \label{tbl:com_scannet_scene}
  \vspace{-3mm}
\end{table*}

\subsection{Object and Scene Reconstruction on Replica}
We begin with the experiments on the Replica dataset, aiming at evaluating the efficacy of our proposed components via a fair comparison of ObjSDF, ObjSDF*, and our method for both scene and object reconstructions in RGB image settings. 
Table~\ref{tbl:com_replica} shows that integrating monocular cues with vanilla ObjectSDF significantly enhances the overall scene reconstruction performance. However, as illustrated in Figure~\ref{fig:ablation}, the reconstructed scene is not entirely smooth and clean. More critically, in the object reconstruction result (\emph{i.e.}, the dining table) in Figure~\ref{fig:ablation}, artifacts appear in the pot area. Our conjecture is that the design of volume-rendering the semantic field in ObjectSDF only ensures the frontal object is in the correct position, but does not guarantee that it absorbs all photons. 
As a result, the geometry of other objects situated behind may cause unexpected artifacts due to light energy leak. 

In contrast, our occlusion-aware object opacity design, which differs from the semantic field rendering in ObjectSDF, results in a considerable improvement, especially in terms of visual quality. As shown in Figure~\ref{fig:ablation}, the artifacts on the dining table geometry are significantly reduced in the result of `Ours w/o reg', which 
also increases the accuracy of the scene geometry by $7.25\%$ in the F-score of the scene reconstruction. However, occlusion-aware object opacity rendering alone is inadequate for eliminating artifacts in the invisible region. Our proposed object distinction regularizer plays an essential role in enhancing the geometry of  invisible regions, leading to further quantitative and qualitative improvements. 
Note that these two components are not computationally intensive and can be easily incorporated into other frameworks.

\begin{figure}
    \centering
    \includegraphics[width=\linewidth]{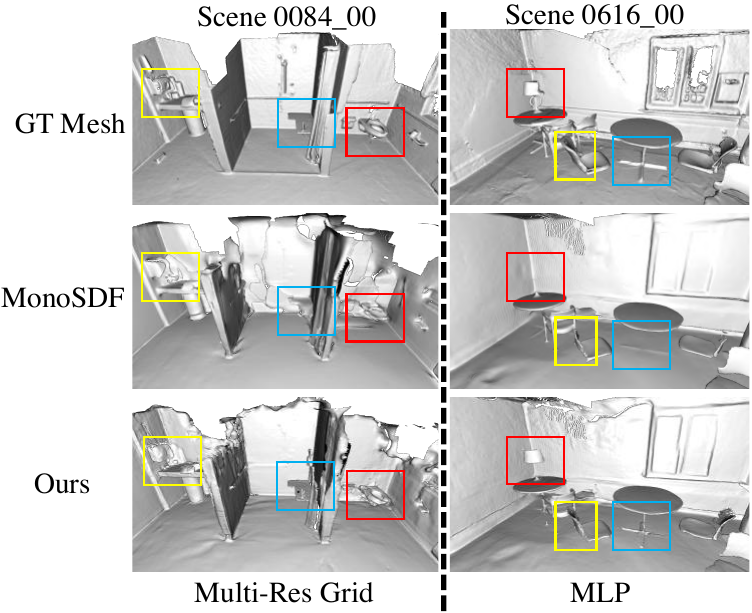}
    \caption{\textbf{Qualitative comparisons on ScanNet}. We compare with the recent MonoSDF~\cite{yu2022monosdf} with two types of network architectures (multi-res grid-based and MLP-based) on  two ScanNet examples. With the help of object-compositional modeling, our method produces better results with more details (\textit{e.g.}, the desk lamp, toilet, and table). 
    All MonoSDF meshes are provided by the original authors. 
    }
\label{fig:scannet_comp}
\vspace{-3mm}
\end{figure}

Figure~\ref{fig:replica_obj_comp} gives more qualitative comparisons of the object reconstruction results on Replica, where we also include the latest RGB-D based object-compositional NeRF method vMAP~\cite{kong2023vmap}.
%
Although vMAP is able to comprehend object surface geometry by utilizing accurate depth information, it still remains vulnerable to imprecise geometry modeling when dealing with occlusion scenarios (e.g., the distortions in the TV screen region) and  geometry within concealed areas (e.g., the chair placed on the wall), as illustrated in Figure~\ref{fig:replica_obj_comp}. In contrast, our proposed method that solely relies on RGB inputs and predicted monocular cues attains impressive geometry reconstruction results by employing our novel occlusion-aware object opacity rendering and object-distinct regularizer.

\subsection{Real-World Large-Scale Scene Reconstruction}
Table~\ref{tbl:com_scannet_scene} gives a comprehensive quantitative comparison with various recent neural implicit surface reconstruction methods which only focus on scene reconstruction without considering the compositing  objects. Even only considering scene reconstruction, our results are either the best or the second best. It is worth noting that MonoSDF reports two different results based on two distinct network structures: multi-resolution (multi-res) grid and multi-layer perceptron (MLP). For a fair comparison, we also implement our model with these two structures. As can be seen from Table~\ref{tbl:com_scannet_scene}, when using the light network architecture (i.e., the multi-res grid structure),  compared to  MonoSDF, our model improves F-score by 10\% and Chamfer distance by 28\%. When employing MLP, 
our quantitative results are comparable to those of the corresponding MonoSDF, while our visual results contain more detailed reconstruction (e.g., the desk lamp in Fig.~\ref{fig:scannet_comp}). Moreover,  
we achieve the best F-score among all methods. 
Note that, unlike MonoSDF, the performance difference of our method with the two different network structures is negligible, which further affirms the robustness of our proposed components.

Fig.~\ref{fig:scannet_comp} gives a qualitative comparison between our method and MonoSDF on ScanNet. No matter which one of the two network architectures is being used, our model produces 
 cleaner surface reconstruction and more details than MonoSDF. This suggests that the overall scene reconstruction quality gets improved when considering the composition of  individual objects in the scene. 

\section{Conclusion}
We have presented \emph{ObjectSDF++}, a new framework for object-composition neural implicit surface reconstruction. Our key observation is that the frontal object should absorb all energy of light during volume rendering. Leveraging this insight, we propose an occlusion-aware object opacity rendering scheme during training. Moreover, we design an object distinction regularization term to penalize the intersection of different objects in  invisible regions. These two techniques significantly improve the quality of  both scene-level and object-level surface reconstruction. Our framework can be extended in the future to support additional types of objects, such as transparent or semi-transparent objects, and be applied to various 3D scene editing applications.

\noindent\textbf{Acknowledgement} This research is partially supported by CRC Building 4.0 Project \#44 and MOE AcRF Tier 1 Grant of Singapore (RG12/22).

\begin{appendices}

\section{Discussion about occlusion-aware object opacity}~\label{sec:dis} We explore the distinctions between our design and other existing methods, providing additional details to enhance the readers' comprehension. Scene opacity is defined as $O_{\Omega}(v)=1-T_{\Omega}(v)$, the probability that a ray does hit a particle before reaching $v$, with rigorously derived PDF $\frac{dO_{\Omega}(v)}{dv}(v)=T_{\Omega}(v)\sigma_{\Omega}(\mathbf{r}(v))$~\cite{yariv2021volume,tagliasacchi2022volume}. 1) When extending to object opacity, a simple way is 
$O_{\mathcal{O}_i}(\mathbf{r})=\int_{v_n}^{v_f}{T_{\mathcal{O}_i}(v)\sigma_{\mathcal{O}_i}(\mathbf{r}(v))}dv, \ (E1)$, which however ignores the occlusions among objects~\cite{yang2021objectnerf,kong2023vmap}. To model the occlusion between objects, vMAP~\cite{kong2023vmap} requires ground truth depth to determine the integration range in~$(E1)$ and requires a manual interval shrinkage design in volume rendering~\cite{oechsle2021unisurf}. This is also similar to the 3D mask design in ObjectNeRF~\cite{yang2021objectnerf} 2) ObjectSDF~\cite{wu2022object} builds an additional semantic field $\mathbf{s}$ and render via 
$\int_{v_n}^{v_f}{T_{\Omega}(v)\sigma_{\Omega}(\mathbf{r}(v))} \mathbf{s}(\mathbf{r}(v)) dv$ to handle occlusion. It introduces an extra tuning hyperparameter for semantic mapping.
3) In contrast, we rethink the occlusion issue and introduce this design: $O_{\mathcal{O}_i}(\mathbf{r})=\int_{v_n}^{v_f}{T_{\Omega}(v)\sigma_{\mathcal{O}_i}(\mathbf{r}(v))}dv$, to approximate both visible and occluded object opacity. This design removes the dependency on ground truth depth data or additional semantic mapping and has shown effectiveness in experiments, which also offers fundamental insights.
\section{More Implementation Details}~\label{sec:app_imp}
\noindent\textbf{Multi-Resolution Feature Grid} Provoked by~\cite{mueller2022instant,tangtorch}, we adopt the multi-resolution feature grid to compensate the fixed frequency position encoding used in vanilla NeRF~\cite{mildenhall2020nerf} to accelerate the model convergence speed. Concretely, the 3D space will be represented by a $L=16$ level of feature grid  with resolution sampled in geometry space to combine different frequencies features:
\begin{equation}
    R_l:= \lfloor R_{min}b^l \rfloor, b:=\exp(\frac{\ln R_{max} - \ln R_{min}}{L-1}),
\end{equation}
where $R_{min}=16, R_{max}=2048$ are the coarsest and finest resolitions, respectively. Each grid includes up to $T$ feature with a dimension of $2$. In the coarse level where $R_l\leq T$, the feature grid is stored densely. For the finer level where $R_l > T$, we follow the Instant-NGP~\cite{mueller2022instant} to apply a spatial hashing function to index the feature vector from the hashing table:
\begin{equation}
    h(x) =(\oplus_{i=1}^3 x_i\pi_i)\text{mod} T
\end{equation}
where $\oplus$ is the bit-wise XOR operation and $\pi_i$ are unique, large prime numbers. The size of the feature vector table $T$ is set as $2^{19}$ similar to~\cite{mueller2022instant,yu2022monosdf}. By concatenating the tri-linear interpolated queried vector from each scale, we append it with the vanilla fixed frequency position embedding of the point coordinates as the input for SDF prediction network~\cite{yu2022monosdf}. 

\noindent\textbf{Geometry initialization for object compositional neural implicit surfaces} To train a model which takes coordinates position as input and then predicts SDF, a good initialization could serve an important role in the optimization. A commonly used technique is the geometry initialization proposed in~\cite{Atzmon_2020_CVPR}. The key design lies in the initial weight to create an SDF field of a sphere in 3D space. In our object-compositional setting, we improve it by manipulating the bias term in the last layer of MLP to create a different radius of the sphere for objects and backgrounds. Specifically, we set the bias term in the channel of common objects as half of that in the channel of background SDF. This design will make sure the objects lie inside the background at the beginning of model optimization. We noticed that this could help in alleviating some object-missing issues during model training. The default radius set for the background object is 0.6-0.9 to cover the camera trajectory. To make sure the minimum operation stays meaningful, we set the region inside the sphere as positive for the background object SDF so that it won't influence the inner object SDF.

\noindent\textbf{Details about Normal and Depth loss} As the monocular depth extracted from pre-trained model~\cite{Eftekhar_2021_ICCV} is not a metric depth, MonoSDF adopts a scale-invariant loss~\cite{eigen2014depth,saxena2005learning} by solving a least-square problem:
\begin{equation}
    (w, q) = \arg \min_{w, q} = \sum_{\mathbf{r}\in \mathcal{R}}(w\hat{D}(\mathbf{r}) + q - \Bar{D}(\mathbf{r}))^2.
\end{equation}
Here the $\Bar{D}(\mathbf{r})$, $\hat{D}(\mathbf{r})$ are the pesudo depth and rendered depth, respectively. This equation has a closed-form solution when sampling larger than 2 points. We solved $w,q$ individually at each iteration for a batch of randomly sampled rays within a single image. The main reason behind it is the depth map predicted by the pre-trained model may differ in scale and shift and the predicted geometry will change at each iteration. Then the depth loss can be defined as:
\begin{equation}
    \mathcal{L}_{depth} = \sum_{\mathbf{r}\in\mathcal{R}}\|w\hat{D}(\mathbf{r}) + q - \Bar{D}(\mathbf{r})\|^2.
\end{equation}

As for the normal loss, we not only force the scale of the normal vector but also the angle similarity for predicted normal and pseudo-normal. The predicted normal vector $\hat{N}$ can also be obtained from the volume rendering result of the gradient of the SDF field, similar to depth and RGB color. The loss can be defined as follow:
\begin{equation}
    \mathcal{L}_{normal} = \sum_{\mathbf{r}\in \mathcal{R}}\|\hat{N}(\mathbf{r}) - \Bar{N}(\mathbf{r})\|_1  + \|1-\hat{N}(\mathbf{r})\Bar{N}(\mathbf{r})\|_1,  
\end{equation}
where $\hat{N}(\mathbf{r})$, $\Bar{N}(\mathbf{r})$ are the rendered normal and pseudo normal from OmniData~\cite{Eftekhar_2021_ICCV} respectively.

\noindent\textbf{Object Distinction Loss} We also provide the idea of the implementation of object distinction regularization loss. Because we only apply this loss to object SDFs which is not the minimum value at this point. We first get the SDF vector $\mathbf{d}(\mathbf{p}) = (d_{\mathcal{O}_1}(\mathbf{p}), d_{\mathcal{O}_2}(\mathbf{p}), \dots, d_{\mathcal{O}_K}(\mathbf{p}))$, then we use the minimum operation to get the scene SDF, $d_\Omega(\mathbf{p})$. We adopt a simple trick to eliminate the influence by subtracting the output of that from the minimum SDF in the loss: 
\begin{equation}
\begin{aligned}
    \sum_{d_{\mathcal{O}_i}(\mathbf{p})\neq d_{\Omega}(\mathbf{p})} & \text{ReLU}(-d_{\mathcal{O}_i}(\mathbf{p})-d_{\Omega}(\mathbf{p}))]\\
    =\sum_{i=1,\dots,K}&[\text{ReLU}(-d_{\mathcal{O}_i}(\mathbf{p})-d_{\Omega}(\mathbf{p}))]\\-&\text{ReLU}(-\min \mathbf{d}(\mathbf{p})-d_{\Omega}(\mathbf{p})),
\end{aligned}
\end{equation}
Due to the minimum operation being differentiable, we are able to calculate this loss and backpropagate the gradient.


\section{Evaluation Metric}~\label{sec:metric}
We provide the definition of the evaluation metrics we used in the main document.
\begin{table}[h]
  \centering
  \begin{tabular}{lc}
    \toprule
    Metric & Definition \\
    \midrule
     Accuracy & $\text{mean}_{\mathbf{p}\in \mathbf{P}}(\min_{\mathbf
{q}\in \mathbf{Q}}\|\mathbf{p}-\mathbf{q}\|_1)$\\
     Completeness & $\text{mean}_{\mathbf{q}\in \mathbf{Q}}(\min_{\mathbf{p}\in \mathbf{P}}\|\mathbf{p}-\mathbf{q}\|_1)$\\
     Chamfer-L1 & 0.5 * (Accuracy + Completeness)\\
     Precision & $\text{mean}_{\mathbf{p}\in \mathbf{P}}(\min_{\mathbf{q}\in \mathbf{Q}}\|\mathbf{p}-\mathbf{q}\|_1) < 0.05$\\
     Recall & $\text{mean}_{\mathbf{q}\in \mathbf{Q}}(\min_{\mathbf{p}\in \mathbf{P}}\|\mathbf{p}-\mathbf{q}\|_1) < 0.05$\\
     F-score & $2*\text{Precision}*\text{Recall}/(\text{Precision+Recall})$\\
     
    \bottomrule
  \end{tabular}
    \caption{\textbf{Evaluation Metric Calculation}. We provide the equation for computing the quantitative metric used in the experiment. Given the sampled point cloud from ground-truth $\mathbf{P}$ and predicted result $\mathbf{Q}$, all the metrics can be calculated as shown above.}
  \label{tbl:com_scannet_scene}
\end{table}

\begin{table*}[t]
  \centering
  \begin{tabular}{l|ccc||cc|cc}
    \toprule
      & \multicolumn{3}{c||}{Model Components} & \multicolumn{2}{c|}{Scene Reconstruction} & \multicolumn{2}{c}{Object Reconstruction} \\
        Method  & Object Guidance & Mono Cue& Regularizer & Chamfer-L1$\downarrow$ & F-score $\uparrow$ & Chamfer-L1$\downarrow$ & F-score $\uparrow$\\
    \midrule
     ObjSDF~\cite{wu2022object} & Semantic &   & & 22.8 & 25.74 & 7.05 & 59.91  \\
     \midrule
     ObjSDF* & Semantic & \checkmark& & 4.14 &78.34& 4.65& 74.06\\
     ObjSDF* w reg & Semantic & \checkmark&\checkmark & 3.96 & 80.58 & 4.18& 76.82 \\
     \midrule
     Ours w/o reg& Occlusion Opacity & \checkmark& & 3.60 & 85.59 & 3.78 & 79.51\\
     Ours & Occlusion Opacity & \checkmark&\checkmark & 3.58 & 85.69 & 3.74 & 80.10 \\ 
    \bottomrule
  \end{tabular}
  \caption{\textbf{The quantitative average results from 8 Replica scenes evaluated on scene and object reconstruction}. We show more results of the ablation study.} 
  \label{tbl:com_replica_supple}
\end{table*}

\section{More Experimental Results}\label{sec:supp_exp}
\subsection{More Details about Experimental Setting}
We use 8 scenes from Replica~\cite{replica19arxiv} following~\cite{yu2022monosdf,kong2023vmap} and 4 scenes from ScanNet~\cite{dai2017scannet} following~\cite{yu2022monosdf,guo2022manhattan} for evaluation. The groud-truth meshes of vMap~\cite{kong2023vmap} are from here\footnote{\url{https://github.com/kxhit/vMAP\#results}} and MonoSDF~\cite{yu2022monosdf} are from here\footnote{\url{https://github.com/autonomousvision/monosdf/blob/main/scripts/download\_meshes.sh}}. vMAP also provides the ground truth object mesh of Replica dataset. We also evaluate our object reconstruction results compared with these data.

\subsection{More results on Replica Dataset}
We also provide the variant of ObjectSDF* with distinction regularization loss in Tab~\ref{tbl:com_replica_supple}. We notice that adding the object distinction regularization loss into ObjectSDF* can further improve the quantitative results. However, the semantic field design limits the surface reconstruction quality, and the quantitative result from `ObjSDF w reg' is still worse than `Ours w/o reg' on the Replica dataset. It demonstrates the effectiveness of occlusion-aware object opacity rendering in improving surface reconstruction quality. We provide more results in Table.~\ref{tbl:replica_chamfer}.

\begin{table*}[t]
  \centering
  \begin{tabular}{lccccccc}
    \toprule
    ~ & Accuracy $\downarrow$& Completeness $\downarrow$& Chamfer-L1 $\downarrow$& Precision $\uparrow$& Recall $\uparrow$& F-Score $\uparrow$ \\
    \midrule
     MonoSDF (Multi-Res Grids)~\cite{yu2022monosdf}& 0.072 & 0.057& 0.064& 0.660&0.601&0.626\\
     \midrule
     ObjectSDF* (Multi-Res Grids) & 0.065 & 0.048 & 0.057 & 0.661 & 0.672 & 0.669 \\
     Ours w/o reg (Multi-Res Grids) & 0.065 & 0.045 & 0.055 & 0.667 & 0.704 & 0.685 \\
     Ours (Multi-Res Grids) & 0.047&0.045&0.046&0.749 &0.707&0.726 \\
    \bottomrule
  \end{tabular}
    \caption{\textbf{The quantitative results of the scene reconstruction  on ScanNet}. We show the ablation results of ObjectSDF++ compared with multi-resolution grid-based MonoSDF. With the introduction of object-compositional modeling, we found the scene reconstruction quality can get significant improvement.}
  \label{tbl:supple_scannet_mono}
  \vspace{-3mm}
\end{table*}

\begin{table}[t]
  \centering
  \begin{tabular}{lcccc}
    \toprule
    ~ & ObjectSDF & ObjectSDF* & ObjectSDF++
    \\
    \midrule
     room0& 17.96/5.26 & 2.76/3.40& 2.68/3.08\\
     \midrule
     room1& 29.29/8.59 & 3.94/5.07 & 3.37/4.66\\
     \midrule
     room2& 28.62/6.15  & 3.30/5.07 &3.03/4.02\\
     \midrule
     office0& 20.10/7.73  & 5.72/4.38 &6.00/3.14\\
     \midrule
     office1& 31.56/11.87 & 6.67/4.50 &4.07/3.79\\
     \midrule
     office2 & 15.98/5.25 & 4.47/4.27 & 3.70/3.62\\
     \midrule
     office3 & 10.29/5.25 & 3.35/5.01 &3.10/3.88\\
     \midrule
     office4 &31.89/10.10 & 2.75/6.53 & 2.72/4.03\\
    \bottomrule
  \end{tabular}
    \caption{\textbf{The quantitative results of the Chamfer distance in individual scenes on Replica, including scene/object.}. We show the individual results from ObjectSDF~\cite{wu2022object}, its variant ObjectSDF* and our framework on different scenes in Replica}
  \label{tbl:replica_chamfer}
\end{table}

\subsection{More results on ScanNet}
We provide more quantitative results of Scannet. The results of ObjectSDF*, Ours w/o reg, and Ours are provided in Tab.~\ref{tbl:supple_scannet_mono}. We found that replacing the semantic field design with the occlusion-aware object opacity training scheme could also show superiority in scene reconstruction quality. The object distinction loss also performs an important role in further improving the quantitative results and making them achieve state-of-the-art performance. It suggests that the combination of object distinction loss and occlusion-aware object opacity rendering scheme is necessary. Besides that, we also find simpling applying the design of ObjectSDF* has already improved the result of MonoSDF (Multi-Res Grid) by a clear margin. It further reassures the benefit of object-compositional modeling in improving surface reconstruction ability.

\subsection{Additional Experimental Results}
To solve the object compositional representation, one simple baseline is to reconstruct each object separately and recombine them together in the final scene. Therefore, we conducted a simple experiment using the same scene as Fig.3.
The results show that independently learning SDF from the mask doesn't correctly give occlusion relationships, resulting in poor reconstruction. Moreover, such a baseline is laborious if there are many objects in a scene.
\begin{figure}[h]
\centering
\includegraphics[width=\linewidth]{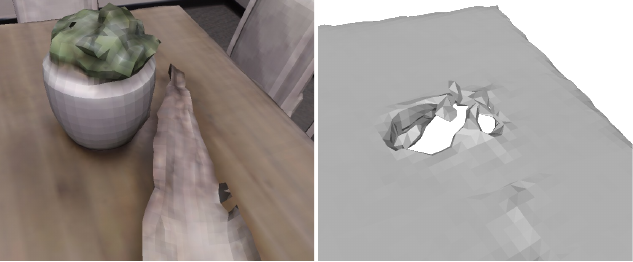}
    \caption{Reconstruct Object separately. Left: reference image, Right: reconstructed desk}
\end{figure}

\section{Limitation about this framework}~\label{sec:discussion}
We improve the quality of ObjectSDF by rethinking the process of opacity rendering and object collision issues. It still exists some space to further improve it. Firstly, although 
 we adopt the multi-resolution grid for accelerating the model convergence speed, the main focus of ObjectSDF++ is not a real-time framework for object-compositional neural implicit surfaces. The estimated training time for one scene is still about 16 hours on Pytorch (depending on how many objects are inside the scene) in a single GPU. 
 We will explore this direction in the future. Secondly, the SDF-based representation is suitable for closed surfaces. It would be better to further extend it to support some open surfaces such as clothes \emph{etc}. Thirdly, the underline assumption of the density transition function is that all objects are solid. Therefore, it is also a good direction to explore whether to represent transparent/semi-transparent objects in the neural implicit surface framework. We also point out that the mask used in this work is a temporally consistent mask. Using an online segmentation mask could enhance the framework's applicability but require additional design for mask association between different frames. We leave it for our future work.
\end{appendices}

{\small
\bibliographystyle{ieee_fullname}
\bibliography{egbib}
}

\end{document}